\documentclass[11pt]{article}

\usepackage[final]{acl}

\usepackage{times}
\usepackage{latexsym}

\usepackage[T1]{fontenc}

\usepackage[utf8]{inputenc}

\usepackage{microtype}

\usepackage{inconsolata}

\usepackage{graphicx}
\usepackage{booktabs}
\usepackage{amsmath}
\usepackage{algorithm}
\usepackage{algpseudocode}
\usepackage{subcaption}

\usepackage{annotates}
\usepackage{mparhack}
\providecommentcommand{rom}{blue}{Roman}
\providecommentcommand{jia}{red}{Jiahui}
\providecommentcommand{sea}{green!30!black}{Sean}

\usepackage{array}
\usepackage{makecell}
\usepackage{tabularx}
\usepackage{booktabs}
\usepackage{multirow}
\usepackage{longtable}
\usepackage{enumitem}

\title{iPOE: Interpretable Prompt Optimization via Explanations}

\author{Jiahui Li$^{1}$, Yarik Menchaca Resendiz$^{1,2}$, Sean Papay$^{1}$ \and Roman Klinger$^{1}$ \\
   $^{1}$Fundamentals of Natural Language Processing, University of Bamberg, Germany \\
   $^{2}$Leibniz-Institut f\"ur Psychologie (ZPID), Trier, Germany \\
   \texttt{\{jiahui.li, sean.papay, roman.klinger\}@uni-bamberg.de} \\
   \texttt{ymr@leibniz-psychology.org}} 

\begin{document}
\maketitle
\begin{abstract}
  Prompt optimization has often been framed as a discrete search
  problem to find high-performing and robust instructions for a large language model (LLM). However, the search result might not make it transparent why
  and where specific prompt changes lead to performance gains. This is
  in contrast to how humans are instructed for annotation tasks. Here,
  researchers carefully design annotation guidelines, leading to
  enhanced annotation consistency. Our paper aims at joining these two
  approaches and introduces iPOE, a novel interpretable prompt
  optimization strategy via explanations. We guide the prompt
  optimization process by automatically created guidelines from
  explanations of annotation decisions (either automatically generated
  or from humans). This set of guidelines is furthermore optimized by
  a series of operations, including removing, adding, shuffling, and
  merging. The resulting prompt includes guidelines that instruct the
  annotation, making the decision process of the LLM and the
  optimization transparent. It therefore supports also laypeople in prompt optimization.  In our experiments on four datasets, we find
  that iPOE can improve over the evaluated baselines by up to $39\%$ and LLM explanations can replace human explanations in the proposed method. Moreover, our interpretability validation study demonstrates that humans and LLMs substantially agree on which guidelines contribute to their annotations, achieving a Cohen’s kappa score between annotators and LLMs of up to $0.68$, $0.80$, and $0.95$ for the emotion classification, medical fact-checking, and hate speech detection tasks respectively.
\end{abstract}
\noindent This paper contains examples with offensive or hateful content.
\section{Introduction}

The effectiveness of large language models (LLMs) often relies on the
expertise of a prompt
engineer~\citep{mishra-etal-2022-cross,psurvey,tan-etal-2024-large}. A
challenge for them is prompt brittleness, wherein minor modifications
in prompt phrasing lead to substantial differences in LLM
outputs~\citep{lu-etal-2022-fantastically,tang-etal-2024-found}. One
approach to support the development of prompts are discrete search
strategies to optimize hard prompts that aim at identifying prompt
variants yielding improved
performance~\citep{pryzant-etal-2023-automatic,prasad-etal-2023-grips}.
However, content-level modifications, such as incorporating explicit
task descriptions or guidance, have a greater impact on model
performance than lexical
perturbations~\citep{lampinen-etal-2022-language,yugeswardeenoo-etal-2024-question,li-etal-2025-humans}.

\begin{figure}[t]
    \centering
    \includegraphics[width=0.95\columnwidth]{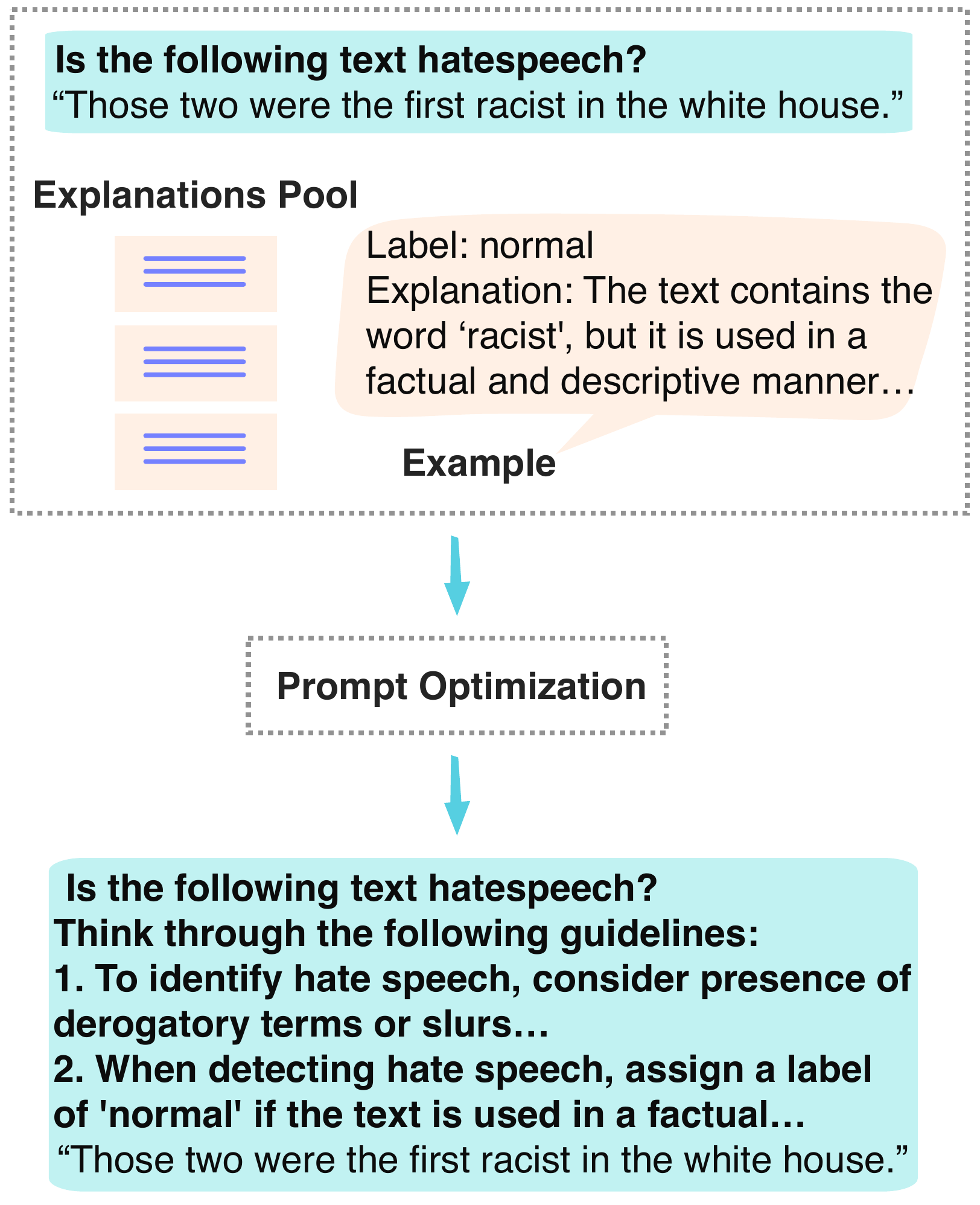}
    \caption{The depiction of our proposed iPOE method to generate
      guidelines as a prompt. The example prompts are bolded and the
      text is quoted.}
    \label{fig:ex}
\end{figure}

A similar strategy for improving human annotator performance is to provide clearer instructions. Such instructions, the
annotation guidelines, are also often iteratively improved, by
measuring inter-annotator
agreement~\citep{griffitt-strassel-2016-query} and adapting them if
needed. Good instructions have a substantial impact on annotation
quality~\citep{Pradhan2022,Sina22}. Interestingly, such annotation
guidelines can also enhance performance by mitigating bias and
ambiguity for LLMs reasoning~\citep{wang2023}.

This combination of using annotation guidelines for humans in LLMs is
arguably a mismatch. As LLMs are not human annotators, applying such guidelines to
a prompt does not guarantee the consistency of LLMs to capture
or adhere to concept definitions specified in these
guidelines~\citep{zhang2023fewshot,fonseca-cohen-2024-large}. This
observation highlights the weakness of applying existing annotation
guidelines, which are designed by humans and targeted at humans,
directly to LLMs. Moreover, manually crafting high-quality guidelines
is itself a challenging task, requiring expertise not only in the task
domain but also in understanding model
behavior~\citep{promptpro,johnnoy}.

We therefore propose to use the idea of iteratively refining
guidelines, as commonly done for humans, but correspondingly with
LLMs. Our approach iPOE (Interpretable Prompt Optimization
via Explanations) derives guidelines from explanations of annotation
decisions. These decision explanations are part of some datasets, but
can also be automatically created posthoc. We illustrate the approach in
Figure~\ref{fig:ex}: The output of our method consists of an augmentation of
an iteratively optimized set of guidelines that extend an existing
prompt for a given task in an interpretable manner. All of our data
and code are available at \url{https://www.uni-bamberg.de/en/nlproc/projects/inprompt/}.

\section{Related Work}

\subsection{Annotation Guidelines for Humans}

Traditional annotation process requires iterative refinement of
guidelines between researchers and annotators to ensure clarity of
tasks~\citep{griffitt-strassel-2016-query}. This process involves multiple rounds of inspecting annotation disagreements, discussing with annotators, and identifying patterns to improve the guidelines. 

The final guidelines often consist of a detailed description of the task, examples, and relevant criteria for particular cases. \citet{Cole11} highlight
the importance of knowledge formulation in guidelines, revealing that
humans' understanding of a task evolves throughout the process of
information searching. The lack of specificity about precise
mechanisms can also bring implicit biases during the annotation
process, leading to invalid experimental designs~\citep{Sina22}.

Although recent crowd-sourcing approaches make annotation more accessible and cost-effective, they can also introduce additional ambiguity. \citet{Pradhan2022} demonstrate that fine-grained annotation guidelines can mitigate such ambiguity and improve data quality. Similarly, \citet{Kulesza14} propose structured labeling solutions to enhance human annotation consistency for machine learning systems. 

\subsection{Prompt Optimization}
%
To address the limitation of manually designed prompts, early work proposed to automatically create prompts for models~\citep{shin-etal-2020-autoprompt}. They formulate prompt optimization as a discrete search problem. Previous methods explored heuristic search, such as beam search for the zero-shot re-ranking process~\citep{cho-etal-2023-discrete}, or the use of edit operations including delete, swap, add, and paraphrase to navigate the prompt space~\citep{prasad-etal-2023-grips}. Similarly, \citet{resendiz-klinger-2023-emotion} propose an automatic prompt optimization approach for emotion-conditioned text generation by adding, removing, or replacing tokens. \citet{yang-etal-2024-dual} and \citet{li-klinger-2025-iprop} search for new prompt candidates by edit-based and LLM rewriting methods supported by human understanding on additional information. Evolutionary strategies were also employed to balance exploration and exploitation in discrete prompt search~\citep{guo2025evopromptconnectingllmsevolutionary}. However, their work lacks contextual information about the task, and the limited search space does not help mitigate potential biases in the task prompt. Therefore, we are inspired to incorporate an expanded context space and frame the problem as searching for an optimal set of guidelines.

Another major research thread leverages preference signals or structured feedback. \citet{pryzant-etal-2023-automatic} introduce natural language “gradients” to critique prompts and guide optimization, while \citet{long-etal-2024-prompt} ask an LLM to propose edits based on the analysis of prompt-output pairs' loss. \citet{cheng-etal-2024-black} refine user prompts by incorporating human and AI preferences, constructing pairs of the original instruction and its optimized version to train a sequence-to-sequence instruction optimizer. Our approach also aims to exploit such additional information either provided by humans or LLMs, but instead to generate rules and guidelines by learning from explanations. 
\begin{figure}[tpb]
    \centering
    \includegraphics[width=0.95\columnwidth]{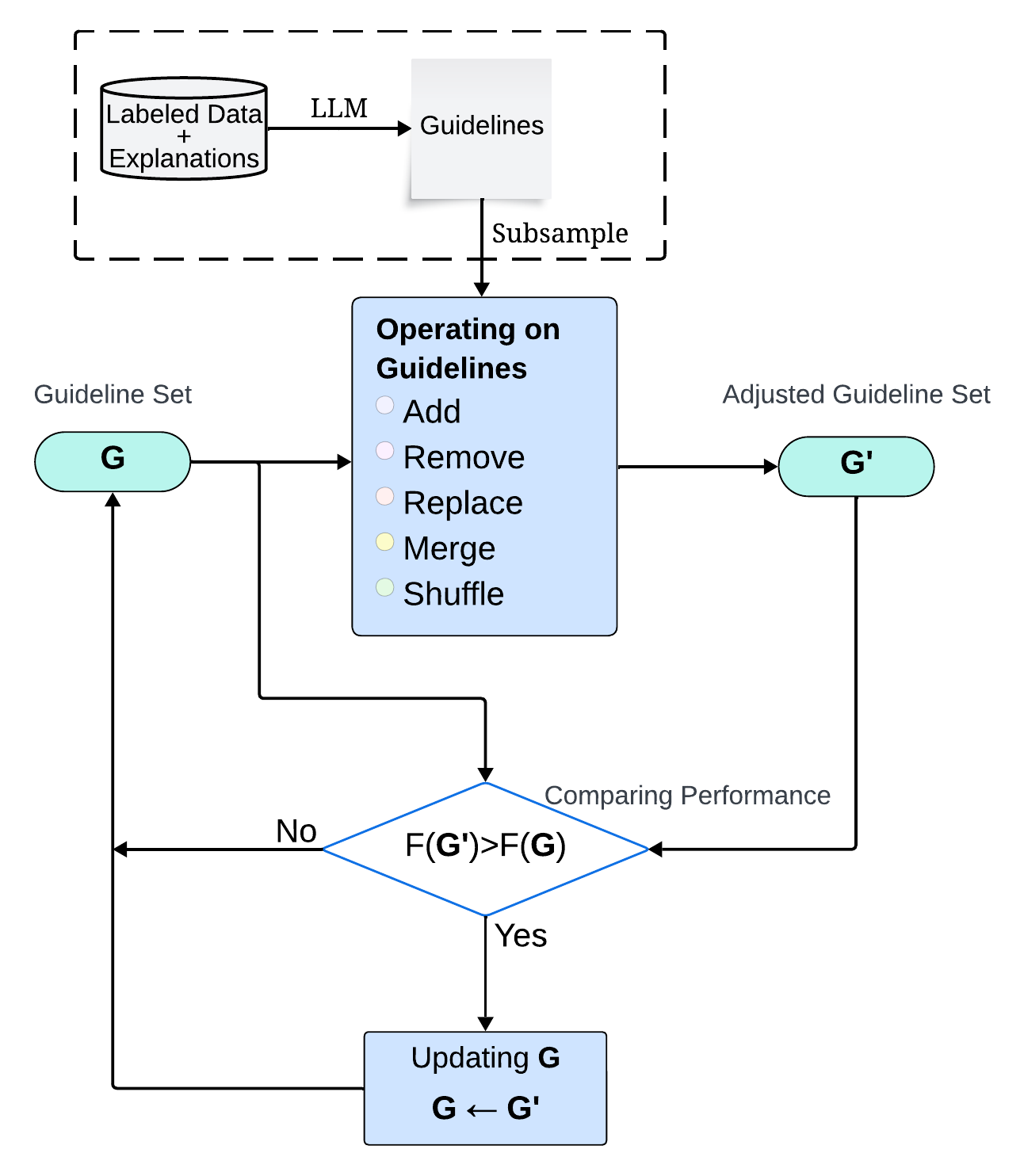}
    \caption{The conceptual workflow of our iPOE approach. $G$ refers to the current guideline set which is an empty set in the very beginning, and $G'$ is the updated guideline set from the operations that achieve the best performance. $F(\cdot)$ refers to a performance metric.}
    \label{fig:opt}
\end{figure}

\subsection{Annotation Guideline Use in Prompts}
Building on top of effective human annotation process, recent work focuses on exploiting annotator guidelines during the model development process. These guidelines are a natural source of information, which constitute high-quality information on the task. \citet{ICLR2024_cda04d7e} fine-tune models to comply with released guidelines by authors, resulting in improved zero-shot information extraction performance. \citet{wang2023} show that expert-written instructions can enhance LLMs on information extraction tasks, emphasizing the value of explicit task guidance. 
In contrast, other studies express less confidence in directly incorporating human annotation guidelines into model inputs. \citet{zhang2023fewshot} report only limited enhancement when integrating annotator guidelines into prompts for LLMs. \citet{fonseca-cohen-2024-large} find that LLMs exhibit inconsistent capabilities when leveraging their manually designed annotation guidelines with conceptual definitions for classification tasks. Thus, our work concentrates on automatically constructing more comprehensive guidelines that adapt to model understanding.

While language models are known to acquire sufficient conceptual structure to generalize across related concepts~\citep{patel2022mapping}, \citet{min-etal-2022-rethinking} observe that LLMs gain limited benefit from input–label examples alone. Therefore, we exploit the generalization abilities of LLMs to generate new annotation guidelines by learning from provided samples along with explanations. 

\begin{algorithm}[tpb]
\caption{Prompt Optimization with Guidelines}
\label{alg}
\begin{algorithmic}[1]
\Require \(P\): prompt prefix; \(D\): training data; \(I\): maximum iterations; \(\mathcal{O}\): guideline operators
\State \(R \leftarrow \mathrm{LLM}(D)\) \Comment{generate guidelines pool}
\State \(G \leftarrow \emptyset\) \Comment{current guideline set}
\State \(S \leftarrow F(P \oplus G)\) \Comment{score of prompt prefix concatenated with current guidelines}
\For{\(i = 1 \ \mathbf{to}\ I\)}
    \State \(g \sim \mathrm{Sample}(R)\) 
    \ForAll{\(o \in \mathcal{O}\)}
        \State \(G' \leftarrow o(G, g)\) 
        \State \(S' \leftarrow F(P \oplus G')\)
        \If{\(S' > S\)}
            \State \(G \leftarrow G'\)
            \State \(S \leftarrow S'\)
        \EndIf
    \EndFor
\EndFor
\State \Return \(P \oplus G\)
\end{algorithmic}
\end{algorithm}
\section{Methods} \label{sec:method}

Figure~\ref{fig:opt} illustrates the conceptual workflow of our iPOE
approach. The labeled data along with its explanations is first converted to a guideline
pool. Then the workflow begins with a prompt prefix and proceeds
through a number of iterations to upgrade the guideline
set. For each iteration, multiple operations are applied with guidelines sampled from the guideline pool. We also provide a formal explanation of the method in Algorithm~\ref{alg}. In this section, we explain the prompt optimization process and
operations used to edit the guideline set in details.

\paragraph{Iterative Optimization.} We first employ an LLM to produce rule-like guidelines from the training data to obtain the guidelines pool $R$ for the optimization process. Each guideline in the pool is generated by prompting the LLM with each instance, gold label, and its explanation on why the label is selected, which are either written by humans or LLMs. The meta-prompts used are presented in Appendix~\ref{sec:metap}.
The workflow begins with a prompt prefix $P$ and the current guideline set $G$ is empty. In each iteration, we subsample a small set of guidelines $g$ and perform a set of guideline operators $\mathcal{O}$ on $g$ respectively. Among all possible operations, we compute the performance scores evaluated on the training data and attain the new guideline set $G'$ that achieves the highest performance:
\[
    G' = \operatorname*{arg\,max}_{\substack{o \in \mathcal{O} \\ g \subset R}} F\!\left(P\oplus o(G, g)\right),
\]
where $F$ denotes the performance metric and $\oplus$ refers to the concatenation of two strings. The framework then checks whether the winning operation yields an improvement:
\[
    F(P\oplus G')>F(P\oplus G).
\]
If the candidate operation improves performance, the guidelines are updated according to
\[
    G \leftarrow G'.
\]
Otherwise, the update is rejected, and the system retains the current guideline set. The above steps are repeated iteratively until a predefined maximum iteration $I$.

\paragraph{Operations.}
Each operation is performed on a set of selected guidelines which is sampled from the generated guidelines database. We provide two sampling strategies: no-control (iPOE) and label-control (iPOE-lb). The no-control method refers to random sampling without any control of which label the sampled guidelines belong to. The label-control method uses uniform sampling according to the gold label of related guidelines, which ensures that guidelines exist in the set for each label. The two methods are employed differently in each iteration when a guideline set is subsampled from the guideline pool, which only affect the `remove' operation.
The full list of operations is:
\begin{itemize}
    \item Add: Add the selected guidelines at the end of the current guideline set.
    \item Remove: Remove $n$ guidelines in the current set randomly for the vanilla method, where $n$ is the number of selected guidelines, or $n=1$ if the current guideline set is smaller than the selected guideline set; remove one random guideline per label that exists in the selected guideline set for the label-control method. 
    \item Replace: Replace the guidelines which are removed in the `Remove' operation with the selected guidelines.
    \item Merge: Merge the selected guidelines with the guidelines in
      the current set that share the same gold label using an LLM. The
      meta-prompts for merging guidelines are shown in
      Appendix~\ref{sec:metap} and an example for this operation is provided in Appendix~\ref{apd:merge}.
    \item Shuffle: Shuffle the order of guidelines in the current set.
\end{itemize}

\section{Experiments} \label{sec:exp}

\begin{table}[tpb]
\centering
\small
\begin{tabular}{lp{0.71\linewidth}}
\toprule
\textbf{Field} & \textbf{Text} \\
\midrule
Instance    & Text: I felt ... when I had to do the dishes earlier. Label: boredom  \\

Exp-h   & I hate the repetitive nature of dishes. Ugh.  \\

Gl-h    & Label the emotion as 'boredom' when the text expresses a negative reaction to a repetitive, monotonous, or tedious task, such as doing dishes, that is performed without engagement or enjoyment, and when the speaker explicitly or implicitly conveys frustration or disdain toward the routine nature of the activity.  \\

Exp-llm & The label 'boredom' was chosen because the text describes a mundane and routine task (doing the dishes) without any strong emotional connotation or excitement. This suggests a lack of engagement or interest, which is characteristic of boredom. \\

Gl-llm  & When classifying emotions in text, select the label 'boredom' if the text describes a mundane or routine task with a lack of strong emotional connotation or excitement, suggesting a lack of engagement or interest.  \\
\bottomrule
\end{tabular}
\caption{An example for human-written and LLM-generated explanations and corresponding guidelines. \textit{Instance}: An instance text and gold label from the crowd-enVent dataset; \textit{Exp-h}: Human-written explanations; \textit{Exp-llm}: LLM-generated explanations; \textit{Gl-h}: Guidelines produced from \textit{Exp-h}; \textit{Gl-llm}: Guidelines produced from \textit{Exp-llm}.}
\label{tab:ex_gl}
\end{table}

We evaluate our iPOE approach on four tasks across three LLMs.
The experiments involve both human-written and LLM-generated
explanations to validate the feasibility of replacing human effort with LLMs for producing guidelines in the iPOE method. Table~\ref{tab:ex_gl} provides an
example of these two explanation types and respective guidelines. The evaluation covers two types of explanations: natural language explanations and feature attributions, which are among the primary explanation types collected and studied in explainability research. The optimization process is conducted for five rounds in total per task to ensure the robustness and stability of the results. In this section, we will discuss the experimental settings and results in detail.

\subsection{Experimental Setting} \label{sec:exp1}
In this section, we first describe the datasets and tasks used in our study, followed by the specific settings employed to obtain LLM predictions. We then outline the baselines used to compare with the iPOE methods. To validate the interpretability of our approach, we further conduct a human study on the generated guidelines, compared with a SHAP~\citep{shap} based explainability method applied to LLMs.

\begin{table}[tpb]
\centering
\begin{tabular}{lcccc}
\toprule
\textbf{Dataset} & \textbf{Train} & \textbf{Va/Te} & \textbf{Exph} & \textbf{Typ}\\
\midrule
crowd-enVent & 5184 & 648 & 985 & Na\\
HealthFC & 600 & 75 & 600 & Na\\
HealthVer & 11464 & 1433 & --- & ---\\
HateXplain & 9132 & 1142 & 9132 & Fe\\
\bottomrule
\end{tabular}
\caption{Statistics of the evaluated datasets. \textit{Train,
    Va/Te}: Number of retained instances for train, validation, and test splits. The validation and test splits contain the same number of instances. \textit{Exph}: Number of available instances that are annotated with explanations; \textit{Typ}: Type of explanation; \textit{Na}: Natural language explanation; \textit{Fe}: Feature attributions explanation.}
\label{tab:dataset_stats}
\end{table}

\paragraph{Tasks.}
We evaluate our approach on four English-language datasets from three
challenging domains in natural language processing: \textsc{crowd-enVent}
dataset~\citep{troiano-etal-2023-dimensional} for emotion
classification,
\textsc{HealthFC}~\citep{vladika-etal-2024-healthfc} and
\textsc{HealthVer}~\citep{sarrouti-etal-2021-evidence-based} datasets for health fact-checking, and \textsc{HateXplain} for hate speech detection~\citep{Mathew2021}. The datasets crowd-enVent and HealthFC provide
explanations in natural language, while HateXplain offers feature attributions as explanations. There are no human-written explanations for the labeled data in HealthVer. All the datasets are split into train, validation, and test sets in a proportion of 80\%, 10\%, 10\% respectively. Their statistics are shown in Table~\ref{tab:dataset_stats}. 

For each dataset, we use human-written and LLM-generated explanations for producing guidelines respectively. The explanations generated by LLMs share the same type with corresponding human annotation. Considering that \textsc{HealthVer} offers no human explanations but shares the same task domain with \textsc{HealthFC}, we use guidelines derived from the human annotation of \textsc{HealthFC} during the optimization process for \textsc{HealthVer}. 

\begin{table}[tpb]
\centering
\begin{tabularx}{\columnwidth}{lXXX}
\toprule
\textbf{Dataset} & \textbf{Qwen3-4B} & \textbf{Llama3-8B} & \textbf{Qwen3-30B} \\
\midrule
crowd-enVent & 296 & 592 & 1777\\
HealthFC & 171 & 343 & 1028\\
HealthVer & 327 & 655 & 1964 \\
HateXplain & 261 & 521 & 1564\\
\bottomrule
\end{tabularx}
\caption{Computational cost of our iPOE methods for three LLMs under evaluated datasets. We report the time estimation (GPU hours) for one whole optimization process including $300$ iterations, which we state as one evaluation round.}
\label{tab:comp_cost}
\end{table}

\paragraph{LLMs.}
We employ three LLMs capable of local deployment:
Llama-3.1-8B-Instruct~\citep{llama3},
Qwen3-4B-Instruct-2507 and
Qwen3-30B-A3B-Instruct-2507~\citep{qwen3technicalreport}.\footnote{\url{https://huggingface.co/}:
    \texttt{meta-llama/Llama-3.1-8B-Instruct}, 
    \texttt{Qwen/Qwen3-4B-Instruct-2507},
    \texttt{Qwen/Qwen3-30B-A3B-Instruct-2507}.
  }
All models are accessed and executed using the HuggingFace library
ecosystem. In order to invoke the creativity of LLM generations,
we enable stochastic decoding via sampling for generating explanations and guidelines.
Specifically, the decoding configuration was selected to match typical
deployments, with \texttt{top\_p=0.9}, \texttt{temperature=0.6}, and
\texttt{top\_k=50}. But the greedy decoding is used for predicting labels to avoid the problem of LLM uncertainty. 
Model predictions are elicited in a zero-shot fashion: for each
instance, the model is presented with the prompt, followed by the
instance text. The generated text is then parsed by attempting to
extract one of the requested labels from a \textit{.json} format
output. The meta prompts used in our study are listed in
Appendix~\ref{sec:metap}. The guideline prompt is evaluated with
F1-macro measure for both training iterations and final testing. The guideline pool is generated from the training splits for each dataset. Considering the computational constraints, the performance score in each iteration is computed on the training splits in a proportion of $0.2$, $1.0$, $0.1$, and $0.1$ for the crowd-enVent, HealthFC, HealthVer, and HateXplain datasets respectively. We set up the randomly sampled guideline number as $3$ per iteration and the maximum optimization iteration as $300$ for each task per evaluation round. The experiments are run on
Nvidia A40 GPUs, with a total estimation of $9,500$ GPU
hours for each round. Table~\ref{tab:comp_cost} reports the computational cost separately for each model and dataset. By comparison, the proposed approach holds comparable number of optimization iterations compared to other prompt optimization methods~\citep{resendiz-klinger-2023-emotion,opro,ape}. 

\begin{table*}[t]
\centering
\small
\setlength{\tabcolsep}{4pt}
\begin{tabular}{llccccccccccc}
\toprule
\textbf{Dataset} & \textbf{Model} 
& \textbf{V} 
& \textbf{R-h} 
& \textbf{R-llm} 
& \textbf{H}
& \textbf{Cot}
& \textbf{IR}
& \textbf{opro}
& \textbf{iPOE-h} 
& \textbf{iPOE-lb-h} 
& \textbf{iPOE-llm} 
& \textbf{iPOE-lb-llm} \\
\midrule

\multirow{4}{*}{crowd-enVent}
 & Qwen3-4B & .35 & .36 & .43 & .35 & .34 & .32 & .35 & \textbf{.50} & .49 & \textbf{.50} & \textbf{.50} \\
 & Llama3-8B & .34 & .43 & .37 & .30 & .35 & .37 & .38 & .48 & \textbf{.51} & .49 & .48 \\
 & Qwen3-30B & .40 & .40 &.45 & .36 & .40 & .39 & .41 & .51 & .51 & \textbf{.54} & .50 \\
\midrule

\multirow{4}{*}{HealthFC}
 & Qwen3-4B & .60 & .46 & .65 & .62 & .62 & .67 & .69 & \textbf{.84} & .83 & .80 & \textbf{.84} \\
 & Llama3-8B & .48 & .38 & .63 & .65 & .52 & .75 & .6 & \textbf{.84} & .76 & .79 & .78 \\
 & Qwen3-30B & .71 & .77 & .76 & .73 & .69 & .63 & .67 & \textbf{.80} & .77 & .79 & .79 \\
\midrule

\multirow{4}{*}{HealthVer}
 & Qwen3-4B & .40 & .35 & .25 & .60 & .43 & .58 & .46 & .62 & .63 & \textbf{.64} & .62 \\
 & Llama3-8B & .41 & .34 & .34 & .61 & .44 & .52 & .4 & .60 & .60 & \textbf{.63} & \textbf{.63} \\
 & Qwen3-30B & .63 & .59 & .59 & .63 & .64 & .65 & .58 & .64 & .64 & \textbf{.68} & .67 \\
\midrule

\multirow{5}{*}{HateXplain}
 & & & & & & & & & & & \begin{tabular}{cc}
\textbf{Fe} & \textbf{Na} \\
\cmidrule(lr){1-1}\cmidrule(lr){2-2}
\end{tabular}
& \begin{tabular}{cc}
\textbf{Fe} & \textbf{Na} \\
\cmidrule(lr){1-1}\cmidrule(lr){2-2}
\end{tabular} \\
 & Qwen3-4B & .44 & .40 & .25 & .45 & .44 & .47 & .50 & .53 & .53 & \begin{tabular}{cc}
.53 & \textbf{.54} \end{tabular} & \begin{tabular}{cc}
.52 & .52 \end{tabular} \\
 & Llama3-8B & .36 & .43 & .48 & .28 & .36 & .32 & .44 & \textbf{.55} & .53 & \begin{tabular}{cc}
.54 & .53 \end{tabular} & \begin{tabular}{cc}
.54 & .51 \end{tabular} \\
 & Qwen3-30B & .45 & .48 & .46 & .40 & .42 & .38 & .44 & \textbf{.57} & .56 & \begin{tabular}{cc}
\textbf{.57} & .55 \end{tabular} & \begin{tabular}{cc}
.54 & .56 \end{tabular} \\

\bottomrule
\end{tabular}
\caption{F1 scores of guideline prompts from different optimization methods for three LLMs. \textit{V (vanilla)}: uses only the prefix prompt. \textit{R-h}: uses the prefix prompt and randomly selected guidelines generated from human explanations. \textit{R-llm}: uses the prefix prompt and randomly selected guidelines generated from LLM explanations. \textit{H}: uses human-written guidelines (see Appendix~\ref{apd:hmgls}). \textit{Cot}: uses chain-of-thought prompts~\citep{Cot}. \textit{IR}: uses prompts generated from the instruction rewriting method~\citep{resendiz-klinger-2023-emotion}. \textit{opro}: uses LLMs as optimizers~\citep{opro}. \textit{iPOE-h}: uses the prefix prompt and guidelines generated from the iPOE method without label control using human explanations. \textit{iPOE-lb-h}: uses the prefix prompt and guidelines generated from the iPOE method with label control using human explanations. \textit{iPOE-llm}: uses the prefix prompt and guidelines generated from the iPOE method without label control using LLM explanations. \textit{iPOE-lb-llm}: uses the prefix prompt and guidelines generated from the iPOE method with label control using LLM explanations; \textit{Fe}: uses feature attributions to generate guidelines; \textit{Na}: uses natural language explanations to generate guidelines. The proposed methods are introduced in Section~\ref{sec:method}, LLMs are explained in Section~\ref{sec:exp1}, and the associated meta-prompts are displayed in Appendix~\ref{sec:metap}.}
\label{tab:f1}
\end{table*}

\paragraph{Baselines.} To evaluate the
effectiveness of our optimization process, we compare the proposed
prompt optimization approach against six baselines: (1)
Vanilla, which uses the prefix prompt without guidelines, (2) Random, which uses a randomly selected guideline set whose size equals the number of labels for each task, (3) Human-written guidelines (presented in Appendix~\ref{apd:hmgls}), which are designed for the annotation process, (4) Chain-of-thought prompts, which use \textit{`Let’s think step by step.'} as an instruction~\citep{Cot}, (5) Instruction rewriting, which iteratively paraphrases prompts based on token-level search~\citep{resendiz-klinger-2023-emotion}, and (6) OPRO, which leverages LLMs as optimizers~\citep{opro}.

\paragraph{Explainability study.} We collect a subset of the judgments for human evaluation on which guidelines assist in their annotation. We evaluate on four different variants of the optimized guideline sets with $120$ sampled instances from the crowd-enVent and HateXplain test splits for the emotion classification and hate speech detection tasks respectively. For the health fact-checking task, we employ three variants of the optimized guideline sets on the full test data of the HealthFC dataset. Due to the high cost of human annotation and ethical reviews, the human study was conducted by three of the authors. The average completion time for
each instance is approximately $45$ seconds and the total study time is about $5$ hours for each annotator. We further implement an explainability analysis based on the SHAP~\citep{shap} method to investigate which guidelines contribute to LLM prediction. The guidelines used in the study are selected from the best performance results from our proposed methods.

\subsection{Results} \label{sec:exp2}

In the following, we present the results and analysis of the experiments described above. We begin by summarizing the performance of the iPOE approach compared to the baseline methods. We select the best performance results among the evaluated five optimization rounds for our proposed methods. Next, we examine the effectiveness of the proposed method when human explanations are replaced with explanations generated by LLMs. We then report a case study in medical fact-checking to evaluate whether guidelines derived from one data distribution can be applied to another distribution within the same domain. Additionally, we investigate whether different types of explanations influence the quality of producing the guidelines. Finally, we qualitatively analyze the evolution of the guideline set in one of the tasks.

\begin{table*}[t]
\centering
\small
\begin{tabularx}{\textwidth}{Xccc}
\toprule
\textbf{Text} & \textbf{Gold} & \textbf{Vanilla} & \textbf{iPOE} \\
\midrule
everything is pretentious when you are ignorant trailer trash inspirational& normal & hatespeech & normal\\
miss my ghetto bitches & normal & hatespeech & normal\\
hiring a cleaning lady as a woman is ghetto debate your mom & normal & hatespeech & normal\\
looks like your average ghetto hopper & normal & offensive & normal\\
we went to the zoo to feed the niglet & hatespeech & normal & hatespeech\\
the muzrat euro marathon & hatespeech & normal & hatespeech\\
why are men actually retarded & offensive & hatespeech & offensive\\
can not fight for shit lmao literally arabs can not fight for shit & offensive & hatespeech & offensive\\
they must be a big minority in d.c. seeing <number> voted for the devil bitch & offensive & hatespeech & offensive\\
thank god no more of that chinaman and protea favourite tabraiz shamsi & offensive & hatespeech & offensive\\
\bottomrule
\end{tabularx}
\caption{Example instances in which the vanilla baseline fails but the iPOE method succeeds. The instances are selected from the HateXplain dataset. \textit{Gold}: the gold label from the original dataset; \textit{Vanilla}: the label predicted by the vanilla baseline; \textit{iPOE}: the label predicted by the iPOE method. The vanilla baseline uses only the prefix prompt without guidelines. For these examples, the iPOE method incorporates human explanations and the Llama-3.1-8B-Instruct model. The prefix prompt and guidelines are provided in Appendix~\ref{apd:comprompt}.}
\label{tab:10ex}
\end{table*}
\paragraph{Overall results.}
Table~\ref{tab:f1} reports the overall F1 scores on the test data. For our iPOE methods, we select the prompts which achieve the highest score on the validation data. We also provide the training and validation learning curves during the prompt optimization process in Appendix~\ref{app:figure}. Across all evaluated datasets and LLMs, our iPOE approach consistently outperforms the baselines. On average, iPOE-h (human explanations without label control) improves over vanilla, random, chain-of-thought, human-written guidelines, instruction rewriting, and OPRO baselines by $34$, $39$, $32$, $25$, $24$, and $23$ percentage points (pp), respectively. iPOE-lb-h (human explanations with label control) outperforms the six baselines by $32$, $37$, $30$, $23$, $22$, and $21$ pp, respectively. These results suggest that label control during the guideline subsampling process has only a minor impact on the general performance. However, we do not view iPOE as competing or conflicting with existing automatic prompt optimization methods. Our proposed approach can be combined with other search-based prompt optimization methods. For example, instruction-rewriting ones, iPOE can benefit from polishing for more concise and minor redundant guidelines. Considering gradient descent ones or optimization based on objective functions, such as OPRO, we can incorporate their underlying principles into our optimization algorithm.

\paragraph{Human explanations vs.\ LLM explanations.}
To reduce human effort for the iPOE approach, we construct guidelines from LLM-generated explanations. As shown in Table~\ref{tab:f1}, iPOE-llm (LLM explanations without label control) improves over vanilla, random (using LLM explanations), chain-of-thought, human-written guidelines, instruction rewriting, and OPRO baselines by $34$, $32$, $33$, $25$, $24$, and $23$ pp, respectively, while iPOE-lb-llm (LLM explanations with label control) achieves gains of $33$, $31$, $31$, $24$, $22$, and $21$ pp, respectively. This indicates that LLM-generated explanations can produce guidelines for the iPOE approach with performance comparable to those derived from human-written explanations. Qualitatively, we observe that guidelines derived from human explanations exhibit similar patterns and logical rules to those generated from LLM explanations. An example of emotion classification task is illustrated in Table~\ref{tab:ex_gl}.

\paragraph{Out-of-distribution evaluation.}
The datasets containing human-written explanations for health fact-checking are scarce, and the selected HealthFC dataset is relatively small. To strengthen the robustness of our evaluation for the health fact-checking task, we specifically apply guidelines derived from the HealthFC dataset to the HealthVer dataset, which does not provide human explanations. Although both datasets address the same task domain, they differ in data distributions. The HealthFC dataset primarily focuses on German clinical trial claims, whereas the HealthVer dataset centers on health-related claims about COVID-19. Our results show that guidelines generated using the iPOE method can effectively support out-of-distribution reasoning within the same domain. Moreover, the out-of-distribution guidelines exhibit capabilities comparable to those of the guidelines generated from the original distribution of the HealthVer training data using LLM explanations. Both sets of guidelines share similar patterns. However, the HealthVer guidelines frequently include examples related to COVID-19.

\paragraph{Feature attributions vs.\ natural language explanations.} 
For hate speech detection, we compare the application of guidelines derived from two explanation modalities including feature attributions and natural language explanations. In our evaluations on emotion classification and health fact-checking tasks, LLM-generated explanations were shown to provide comparable support for guideline generation. We therefore use LLMs to generate natural language explanations from the labeled data in the HateXplain dataset, which originally provides only feature attribution explanations. Our results indicate that both explanation modalities yield guidelines with similar patterns and comparable effectiveness during the optimization process.

\paragraph{Qualitative analysis.}

To further compare the difference between the iPOE approach and the baselines, we present several examples along with the predictions of the iPOE-h and vanilla methods in Table~\ref{tab:10ex}. The associated prompts are provided in Appendix~\ref{apd:comprompt}. For the hate speech detection task, we specifically select $10$ instances in which the vanilla method fails but the iPOE method succeeds. The examples illustrate incorrect predictions generated by the vanilla method from three perspectives: (1) instances labeled as `normal' are misclassified as `hatespeech' or `offensive' text; (2) instances labeled as `hatespeech' are predicted as `normal'; and (3) instances labeled as `offensive' are wrongly viewed as `hatespeech'.
In case (1), the guidelines provide clarifications for the label `normal' and include examples of expressions that may appear ironic but are not hateful. In case (2), the guidelines list typical derogatory terms associated with particular groups or individuals to identify genuinely hateful content. In case (3), the guidelines present specific rules and examples that help distinguish offensive content from hate speech.
Across all cases, the vanilla prompt lacks such clarifications and examples, which likely contributes to the model’s misunderstandings and resulting misclassifications.

Additionally, we provide an example of the iPOE development process trained on the HealthVer dataset in Appendix~\ref{apd:gls_dev}. We observe that the guidelines produced by LLMs tend to be longer than the corresponding explanations. The LLMs often supplement the guidelines with a formal description of how to select a specific label, along with typical instances and illustrative examples. We agree that this additional information helps clarify the task and improves readability. In addition, some guidelines not only describe the features of the selected label but also explain how it differs from other labels, which assists in reducing bias and ambiguity in the task. To investigate the effect of incorporating richer guidelines during iterative search, we analyze the number of guidelines included in the optimal prompts. We observe that cases in which the number of generated guidelines exceeds the size of the label set are rare. 
By plotting the frequency of each operation in the optimization process, we find that add, remove, and replace operations yield comparable contributions, while the merge operation accounts for the largest proportion of edits across all tasks. 
This operation rewrites multiple guidelines derived from the same gold label into a more comprehensive yet concise form. Since the guidelines are generated by LLMs and the merge operation is also performed by the same LLM, we do not conduct additional experiments that remove the merge operation or replace it with simple concatenation or summarization, primarily due to the associated computational cost. Moreover, previous studies have shown that prompt brittleness in recent LLMs is relatively insensitive to semantically equivalent modifications~\citep{li-etal-2025-humans}. Based on a careful qualitative review, we observe that the guidelines before and after the merge operation are not fundamentally different, suggesting that the performance gains are only marginally influenced by LLM-based rewriting.

\begin{table}[tpb]
\centering
\begin{tabularx}{\columnwidth}{lXXX}
\toprule
\textbf{Dataset} & \textbf{Qwen3-4B} & \textbf{Llama3-8B} & \textbf{Qwen3-30B} \\
\midrule
crowd-enVent & .63 & .68 & .56\\
HealthFC & .65 & .80 & .74\\
HateXplain & .83 & .81 & .95\\
\bottomrule
\end{tabularx}
\caption{Average Cohen’s kappa scores between humans and LLMs on selected guidelines of the evaluated datasets for our explainability study.}
\label{tab:kappa}
\end{table}

\paragraph{Interpretability validation.} We compare which guidelines contribute to annotation decisions between humans and LLMs, and examine their inter-annotator agreement. We observe that humans tend to focus on guidelines directly associated with their selected label. In contrast, SHAP explanations indicate that multiple guidelines can contribute to the LLM’s predictions. We hypothesize that humans may primarily attend to guidelines supporting the chosen label while ignoring those that help rule out alternative labels, whereas LLMs incorporate both types of information. Overall, we obtain Cohen’s kappa agreement scores of selected guidelines between humans and LLMs for the evaluated tasks and LLMs, presented in Table~\ref{tab:kappa}. The kappa scores indicate substantial agreement for the emotion classification, health fact-checking, and hate speech detection tasks, thereby providing evidence to some extent that humans partially parallel their decision-making process with the model's.

\section{Conclusion and Future Work}
We proposed interpretable prompt optimization as a novel approach to produce guidelines using either human-written or LLM-generated explanations. The guidelines are expected to support LLMs' prediction. To search for the optimal guideline set, we further iteratively apply a series of operations including removing, adding, shuffling, and merging. With this approach,
we instruct the LLM's annotation by effective guidelines, offering a transparent decision process of the LLM and the optimization.

Our results span three domains and suggest that iPOE can enhance performance while producing prompts that are human-understandable and interpretable. We demonstrate that LLM-generated explanations can effectively replace human-written explanations within the proposed approach, thereby reducing the need for costly human effort. Moreover, the optimized guideline set achieves higher performance than human-written annotation guidelines in LLM prediction. The out-of-distribution evaluation also provides encouraging evidence that iPOE can be applied to other domains where human annotations are difficult to obtain.

The proposed approach has revealed several challenges that deserve further investigation. There is
a need to explore more efficient methodologies for a faster convergence of the learning process. Additionally, some optimal guideline sets contain duplicate guidelines associated with the same label. The readability of our approach could be improved by incorporating methods that effectively identify and eliminate such redundancies.

\section*{Limitations}

Despite the promising performance of our approach, we observe several limitations. First, the availability of human-written explanations, especially feature attributions, is limited. Consequently, the comparison between human-generated and LLM-generated explanations, as well as across different explanation types, remains constrained in scope. A more comprehensive analysis would require a more diverse collection of human annotations. Second, due to computational constraints, we do not evaluate the impact of varying numbers of sampled guidelines in each optimization iteration to measure the sensitivity of the proposed method to the sample size. Third, although the guideline pool generated from human explanations is smaller than that generated from LLM explanations while achieving comparable performance, further ablation studies are needed to investigate the effect of guideline pool size under the same source conditions. Lastly, a larger scale human study would provide stronger validation of our interpretability claims, although such studies are limited by associated costs.

\section*{Ethical Considerations}
The datasets used for our experiments are publicly available and we ensure that they have been collected according to ethical standards before employing them. We do not change the intentional use for the evaluated datasets and LLMs. Our method does not contribute to the republication or redistribution of any datasets and models. We believe our findings provide valuable insights for future research in automatic annotation processes using LLMs. We acknowledge that the evaluated datasets involve potential ethical challenges. However, as our work does not introduce a novel task and instead builds upon existing tasks, we do not anticipate additional ethical concerns beyond those already associated with prior work. The human study is conducted by the authors of this paper. Besides, the paper presents examples including offensive content or hate speech, which we give a warning in the beginning. Additionally, ChatGPT~\citep{chatgpt} was used to improve code generation for making figures and tables, as well as to assist with grammar and vocabulary in the text of some paragraphs of this paper.

\section*{Acknowledgments}

This work is funded by the project INPROMPT (Interactive Prompt
Optimization with the Human in the Loop for Natural Language
Understanding Model Development and Intervention, funded by the German Research Foundation, KL 2869/13--1, project number 521755488).
We also gratefully acknowledge the scientific support and HPC resources provided by the Erlangen National High Performance Computing Center of the Friedrich-Alexander-Universität Erlangen-Nürnberg under the BayernKI project. BayernKI project funding is provided by Bavarian state authorities.
\bibliography{custom}

\appendix

\section{Appendix}
\label{sec:appendix}

\subsection{Meta prompts} \label{sec:metap}
Table~\ref{tab:metap} presents our manually designed meta prompts for generating explanations and guidelines. In our prompt optimization process, we also use a structured system prompt that enforces LLMs to output text in a \textit{.json} format. 

\begin{table*}[t]
\centering
\small
\setlength{\tabcolsep}{6pt}
\renewcommand{\arraystretch}{1.15}

\begin{tabularx}{\linewidth}{
>{\raggedright\arraybackslash}p{0.28\linewidth}|
>{\raggedright\arraybackslash}X}
\toprule
\textbf{Type} & \textbf{Prompt} \\
\midrule

Natural language explanation &
\begin{minipage}[t]{\linewidth}\ttfamily
"role": "user", "content":\\
Provide a brief explanation for why the given label was chosen in the above task. When writing the explanation, you may describe the key cause or feature that led to your decision, link it to the general rule, principle, or pattern defined in the task, and keep only relevant details.\\
Text: \{text\}\\
Label: \{label\}\\
\end{minipage}
\\
\midrule

Feature attributions explanation &
\begin{minipage}[t]{\linewidth}\ttfamily
"role": "user", "content":\\
Highlight the words or phrases in the text that contribute most significantly to the assignment of the given label.\\
Text: \{text\}\\
Label: \{label\}\\
\end{minipage}
\\
\midrule

Guideline for natural language explanation &
\begin{minipage}[t]{\linewidth}\ttfamily
"role": "user", "content":\\
Using the provided sample text and its corresponding human annotation, along with a list of feature attribution explanations that are most responsible for why the label was chosen, provide a rule-based guideline for performing this \{task\_name\} task. The guideline should be written in one paragraph. \\
Text: \{text\}\\
Label: \{label\}\\
Explanation: \{explanation\}
\end{minipage}
\\
\midrule

Guideline for feature attributions explanation &
\begin{minipage}[t]{\linewidth}\ttfamily
"role": "user", "content":\\
Using the provided sample text and its corresponding human annotation, along with a list of feature attribution explanations that are most responsible for why the label was chosen, provide a rule-based guideline for performing this \{task\_name\} task. The guideline should be written in one paragraph. \\
Text: \{text\}\\
Label: \{label\}\\
Explanation: \{explanation\}
\end{minipage}
\\
\midrule

Merge guidelines &
\begin{minipage}[t]{\linewidth}\ttfamily
"role": "user", "content":\\
Please rewrite the following guidelines into one guideline. \\
1: \{guideline 1\}\\
2: \{guideline 2\}\\
...
\end{minipage}
\\
\midrule

Structured system prompt &
\begin{minipage}[t]{\linewidth}\ttfamily
"role": "system", "content":\\
Provide the explanation in exactly one valid JSON object — nothing else. The JSON object must have exactly one field:\\
\{\\
\ \ "\{feature\_name\}": "<\{feature\}>"\\
\}

\end{minipage}
\\

\bottomrule
\end{tabularx}

\caption{Meta prompts to generate different types of explanations and respective guidelines.}
\label{tab:metap}
\end{table*}

\subsection{An example for the `Merge' operation} \label{apd:merge}
Table~\ref{tab:ex_merge} illustrates an example for the `Merge' operation introduced in Section~\ref{sec:method}. This example is derived from the HealthFC dataset using guidelines generated from human-written explanations by the Llama-3.1-8B-Instruct model. The updated guideline set is the result of merging the current guideline set and sample guidelines.
\begin{table*}[t]
\centering
\small
\begin{tabular}{l!{\vrule}p{0.75\linewidth}}
\toprule
\textbf{Field} & \textbf{Text} \\
\midrule
\textbf{Current Guideline Set} 
& 1. Identify the claim being made, which in this case is whether cancer can be detected by dark field analysis of the blood. Determine if the evidence provided supports or refutes the claim; if the evidence is inconclusive, consider any limitations mentioned, such as the number of participants. Evaluate the credibility of the evidence, noting if a thorough search was conducted across multiple research databases and if the results were consistent. Assess the implications of the evidence on the claim, considering the potential consequences of a false diagnosis, such as uncertainty among those affected. Based on the evaluation, assign a label to the claim, such as "\textbf{refuted}" if the evidence indicates that the claim is false or unsupported, and provide a clear explanation for the chosen label. 2. To label a claim as \textbf{supported}, the evidence must provide a well-established and quantifiable benefit of the treatment (e.g., 'an average of four to five days' faster healing) for a specific type of wound or condition (e.g., second-degree burns) that is consistent with the explanation and not contradicted by other information in the text. 3. label the task as "\textbf{not enough information}" if the provided text and its corresponding annotation indicate that the effectiveness or reliability of a medical intervention or measure (such as screening) is uncertain, inconclusive, or lacks solid data, as evident from phrases like "it is currently uncertain," "cannot reliably identify," or "not well established."  \\
\midrule
\textbf{Sampled Guidelines}
& 1. To perform the medical language inference task, label a claim as ""\textbf{not enough information}"" when the evidence provided either lacks direct studies examining the healing effects of the method in question, or when existing evidence contradicts a fundamental assumption of the method, but does not provide conclusive proof of its efficacy or ineffectiveness. 2. Identify the claim and the evidence provided to support or refute it, then assess the quality and reliability of the evidence by looking for high-quality studies and expert opinions. Next, evaluate the key findings of the evidence, focusing on the comparison of the active ingredient rivaroxaban to other anticoagulants, and determine whether the evidence supports or refutes the claim. If the evidence shows that rivaroxaban has a lower risk of bleeding than other anticoagulants, or if there is no significant difference, then the claim is \textbf{refuted}. However, if the evidence suggests that rivaroxaban has a higher risk of bleeding than other anticoagulants, then the claim is supported. 3. Identify the claim being made, which should be a clear and concise statement about a medical topic. Locate the evidence provided to support or refute the claim, which should include references to credible sources, such as research papers. Evaluate the strength and relevance of the evidence, considering factors like the study design, sample size, and outcome measures. Assess the consistency and reliability of the findings, noting any limitations or potential biases in the evidence. Consider alternative explanations for the observed effects, accounting for potential confounding variables and the possibility of chance or coincidence. Determine the scope and applicability of the evidence, recognizing any gaps in the research, particularly regarding specific populations, such as women. Based on the evaluation of the evidence, decide whether the claim is \textbf{supported}, refuted, or remains uncertain, and provide a clear justification for the chosen label.  \\
\midrule
\textbf{Updated Guideline Set}
& 1. To evaluate the claim about a medical topic, follow these steps: (1) identify the claim and locate the provided evidence, including references to credible sources; (2) assess the strength, relevance, and consistency of the evidence, considering factors like study design, sample size, outcome measures, limitations, and potential biases; (3) evaluate the scope and applicability of the evidence, accounting for gaps in the research, particularly regarding specific populations; (4) based on the evaluation, decide whether the claim is supported, \textbf{refuted}, or remains uncertain, and provide a clear justification; and (5) for specific claims about anticoagulants, such as rivaroxaban, assess the key findings and compare the active ingredient to other anticoagulants, refuting the claim if the evidence shows a lower risk of bleeding or no significant difference. 2. To label a claim, identify the claim being made and locate the evidence provided to support or refute it. Evaluate the strength and relevance of the evidence, considering factors like study design, sample size, and outcome measures. Assess the consistency and reliability of the findings, noting limitations and potential biases. Consider alternative explanations and the scope and applicability of the evidence. Determine whether the claim is \textbf{supported}, refuted, or remains uncertain based on the evaluation, and provide a clear justification for the chosen label. A claim is considered supported if the evidence provides a well-established and quantifiable benefit for a specific type of wound or condition that is consistent with the explanation and not contradicted by other information in the text. 3. Label the task as '\textbf{not enough information}' if the provided text and its corresponding annotation indicate that the effectiveness or reliability of a medical intervention or measure is uncertain, inconclusive, or lacks solid data, as evident from phrases like 'it is currently uncertain,' 'cannot reliably identify,' or 'not well established.' This includes situations where the evidence lacks direct studies examining the healing effects of the method in question or when existing evidence contradicts a fundamental assumption of the method, but does not provide conclusive proof of its efficacy or ineffectiveness. \\
\bottomrule
\end{tabular}
\caption{An example for the `Merge' operation from the HealthFC dataset using guidelines generated from human-written explanations by the Llama-3.1-8B-Instruct model. The labels for generating corresponding guidelines are bolded in the text.}
\label{tab:ex_merge}
\end{table*}

\subsection{Learning curves of F1 scores for iPOE} \label{app:figure}
\begin{figure*}[ht]
    \centering
    \includegraphics[width=0.95\linewidth]{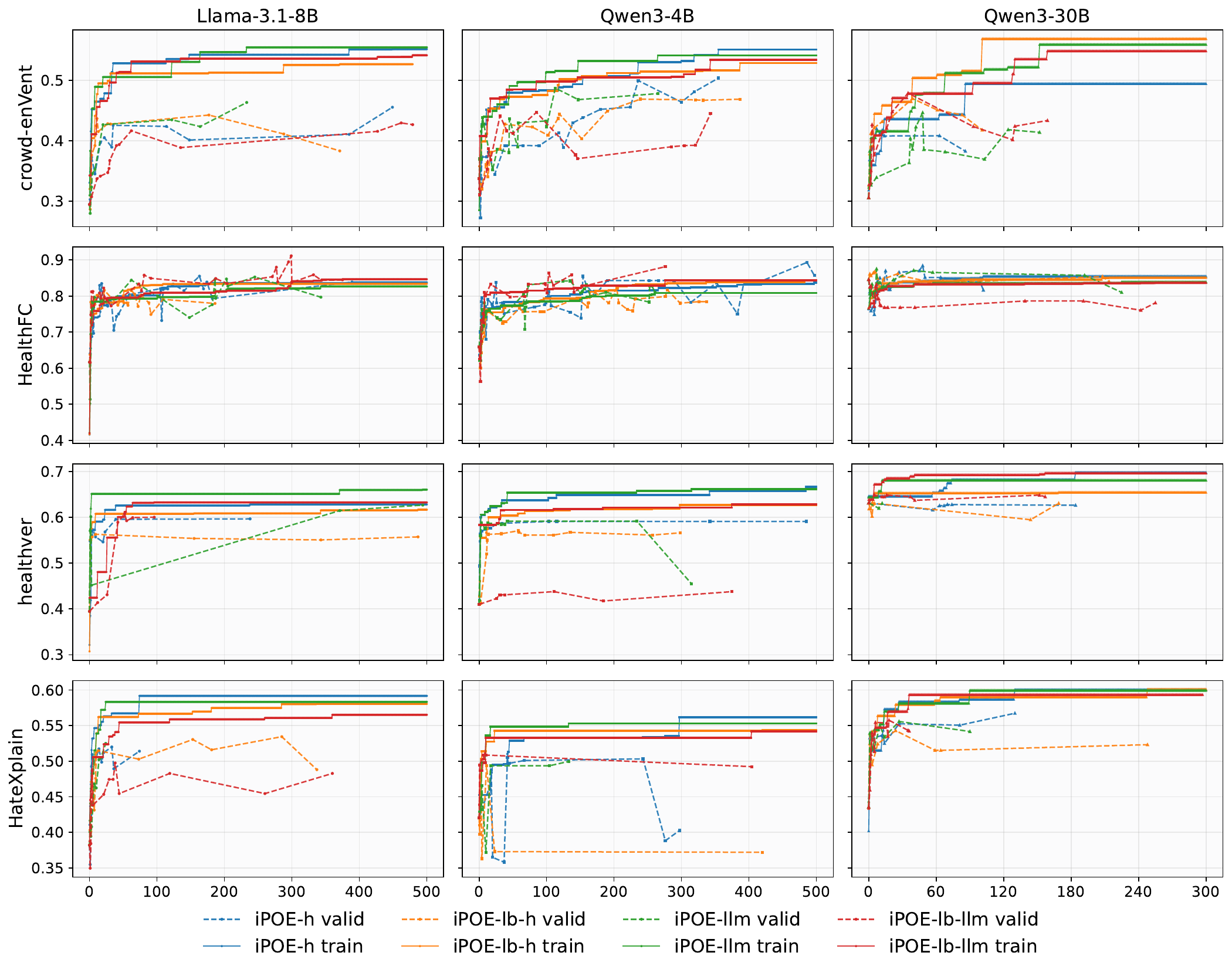}
    \caption{Learning curves of F1 scores on the training and validation sets for our iPOE approach. Each plot corresponds to a (dataset, LLM) pair introduced in Section~\ref{sec:exp1}. LLM names are shown at the top, and dataset names on the left. The legend for each method is displayed at the bottom and is introduced in Section~\ref{sec:method}. \textit{iPOE-h}: use the prefix prompt and guidelines generated from the iPOE method without label control using human explanations. \textit{iPOE-lb-h}: use the prefix prompt and guidelines generated from the iPOE method with label control using human explanations. \textit{iPOE-llm}: use the prefix prompt and guidelines generated from the iPOE method without label control using LLM explanations. \textit{iPOE-lb-llm}: use the prefix prompt and guidelines generated from the iPOE method with label control using LLM explanations.}
    \label{fig:f1-all}
\end{figure*}

Figure~\ref{fig:f1-all} plots the learning curves of F1 scores on the training and validation sets for the proposed approach. The plots are arranged in a $4 \times 3$ grid, corresponding to four datasets and three LLMs.

\subsection{Example prompts for the vanilla and iPOE methods} 
In Section~\ref{sec:exp2}, we discuss $10$ instances for the hate speech detection task where the vanilla baseline fails but our iPOE method succeeds. We provide the prompts used in these two methods in Table~\ref{tab:comprompt}.

\label{apd:comprompt}
\begin{table*}[tpb]
\centering
\small
\begin{tabularx}{\textwidth}{cX}
\toprule
\textbf{Method} & \textbf{Prompt} \\
\midrule
Vanilla & Classify the following text based on whether it contains abusive language using one of the labels hatespeech, offensive, normal.\\
iPOE & Classify the following text based on whether it contains abusive language using one of the labels hatespeech, offensive, normal. Think through the following guidelines before giving the final answer. guidelines: 1. Assign the 'hatespeech' label to a text if it contains a word or phrase that is explicitly derogatory or pejorative towards a particular group or individual, such as racial or ethnic slurs, as indicated by the presence of a feature attribution explanation that identifies a term like 'kike' as the primary reason for the label choice. 2. explicit profanity or slurs, such as words with derogatory connotations towards a specific group, and aggressive or condescending language, including phrases that imply intellectual superiority or belittling of others. The presence of either feature, especially when used in a derogatory manner, is sufficient to warrant the 'offensive' label. 3. When detecting hate speech, assign a label of 'normal' if the text contains a feature attribution that is a pejorative term, but is used to describe a group or individual in a way that is intended to be ironic or critical, rather than hateful or derogatory. For example, if a word like 'ghetto' is used to describe a name or a group in a way that is meant to highlight its perceived authenticity or lack of pretension, rather than to express contempt or disdain, then the text should be labeled as 'normal'. 4. To detect hate speech, identify and flag instances where derogatory terms such as 'illegal', inflammatory labels like 'immigrants', and phrases that convey exclusion or denial of identity, e.g., 'aren't americans', are used. Additionally, pay attention to grammatical irregularities, such as the use of possessive apostrophes in non-standard contexts, denoted by '’'. 5. To detect hate speech, identify and count instances of derogatory feature attributions such as slurs, insults, and ethnic or religious epithets, including but not limited to words with derogatory connotations like 'arsed','moslem', and'shite', and classify the text as hate speech if at least one such attribution is present. 6. Assign the label 'offensive' if the text contains phrases that explicitly condone violence, promote discriminatory ideologies, or dehumanize specific groups, such as the mention of 'rape' and 'abuse' in relation to women, or if it expresses contempt towards entire societies, like 'western society is corrupt' or describes certain groups as 'inferior', and if the text is associated with a specific ideology or institution that is known to promote hate speech, like 'islam has faith schools'. 7. if a text contains words or phrases that explicitly target or demean a specific group based on their race, ethnicity, or religion, such as derogatory terms ('niggers', 'jews') or phrases that assign inferior roles or power ('only men have political power'), and the tone is threatening or aggressive ('obey or leave', 'will volunteer for the ovens'), classify the text as hate speech. 8. To detect hate speech, identify instances where derogatory terms or slurs are used to target specific ethnic or national groups, as indicated by the presence of words like 'pakis', which is a derogatory term for people of Pakistani descent. Consider the context and tone of the text, and if it contains language that is discriminatory, offensive, or incites hatred towards a particular group, classify it as hate speech. 9. Assign a label of 'offensive' if the text contains words or phrases that express hatred or intolerance towards a particular group, such as'muslim' or 'preachers', and the tone is critical or dismissive, as indicated by phrases like 'are free to spout their hatred unhindered'. 10. To identify hate speech, look for the use of derogatory terms such as racial slurs (e.g. 'nigger'), explicit references to racist stereotypes (e.g.'magic negro'), and phrases that promote segregation or exclusion (e.g. 'be somewhere else'). Additionally, consider the context in which these terms are used, and whether they are accompanied by language that implies a desire to exclude or marginalize individuals based on their race or ethnicity. 11. When detecting hate speech, if the text contains words or phrases that are explicitly derogatory or pejorative towards a particular group or individual, such as 'kike','shill', or phrases that reference conspiracy theories like 'the matrix', and the overall tone of the text is hostile or exclusionary, classify the text as hate speech. 12. When detecting hate speech, consider the presence of derogatory terms or slurs that target individuals with disabilities, such as'retard', and label the text as 'offensive' if any of these terms are found, as they are widely recognized as hurtful and demeaning language.\\

\bottomrule
\end{tabularx}
\caption{An example for prompts used in the vanilla and iPOE methods evaluated on the HateXplain dataset. The vanilla baseline is described in Section~\ref{sec:exp1}. The iPOE prompt is created using human explanations by the Llama-3.1-8B-Instruct model, which achieves the highest performance on the validation data. }
\label{tab:comprompt}
\end{table*}

\subsection{Human-written Guidelines} 
In Section~\ref{sec:exp2}, we compare our proposed iPOE methods with multiple baselines. We provide the human-written guidelines used as one of the baselines in Table~\ref{tab:hmgls}.

\label{apd:hmgls}
\begin{table*}[tpb]
\centering
\small
\begin{tabularx}{\textwidth}{cX}
\toprule
\textbf{Task} & \textbf{Guidelines} \\
\midrule
Emotion Classification & Put yourself in the shoes of other people. The text
describes events that occurred in the life of their authors. Don’t be surprised if
they are not perfectly grammatical, or if you find that some words are missing.
For each event, you will assess if it provoked an emotion in the experiencer,
and if so, what emotion that was. Moreover, you will be asked how you think
the experiencer assessed the event: you will read some statements and indicate
how much you agree with each of them on a scale from 1 to 5. The writers of
these texts have answered these questions in a previous survey. Your goal now
is to guess the answer given by the writers as closely as possible. How confident are you about your answer (1-5)?  How long do you think the event lasted (seconds; minutes; hours; days; weeks)?  How long do you think the emotion lasted if the experiencer had any (seconds; minutes; hours; days; weeks; this event did not cause any emotion)? How intense do you think the emotion was? \\
Health Fact Checking & The label NOT ENOUGH INFORMATION is used if the evidence is neutral or irrelevant to the claim. The evidence statement could be complete or incomplete. For the claims that include multiple pieces of information, the SUPPORTED label is considered if the evidence supports one of them and the other pieces of information were neutral. Similarly, the REFUTED label is considered if the evidence refutes one of the multiple pieces of information. For
neutral (NOT ENOUGH INFORMATION) examples, we encourage you to select sentences that are relevant but do not contain enough information to make a decision. \\
Hate Speech Detection & Hate speech is considered any kind of content that conveys
malevolent intentions toward a group or an individual, and motivated by inherent characteristics that are attributed to that group and shared among its members such as race, color, ethnicity,
gender, sexual orientation, nationality, religion, disability, social
status, health conditions, or other characteristics. The outcome of Hate Speech could be the
promotion of division among people, undermining of social cohesion in communities, inciting others to commit violence or
discrimination, and could have consequences for individuals’
health and safety. However, even if it is offensive, it is not considered Hate Speech any content that attacks a person’s personality traits, ideas, or opinions. Hate Speech can also be implicit, portrayed as
an indirect or coded language that uses Irony, Stereotypes, or
Misinformation. \\

\bottomrule
\end{tabularx}
\caption{Human-written guidelines used as a baseline. For the emotion classification task, we use the original annotation guidelines provided in the crowd-enVent dataset~\citep{Troiano2023}. The guidelines for the health fact-checking task are adapted from the HealthVer dataset paper~\citep{sarrouti-etal-2021-evidence-based}. For the hate speech detection task, we employ annotation guidelines derived from expert definitions~\citep{melis-etal-2025-modular}.}
\label{tab:hmgls}
\end{table*}

\subsection{An example for guidelines development} \label{apd:gls_dev}
We present an example for the iterative guidelines development process trained on the HealthVer dataset and the Qwen3-4B-Instruct-2507 model using the iPOE-llm method in table~\ref{tab:gls}. 

\subsection{Instructions of Survey for Human Study}
\begin{figure*}[htbp]
\centering
  \includegraphics[width=0.8\linewidth]{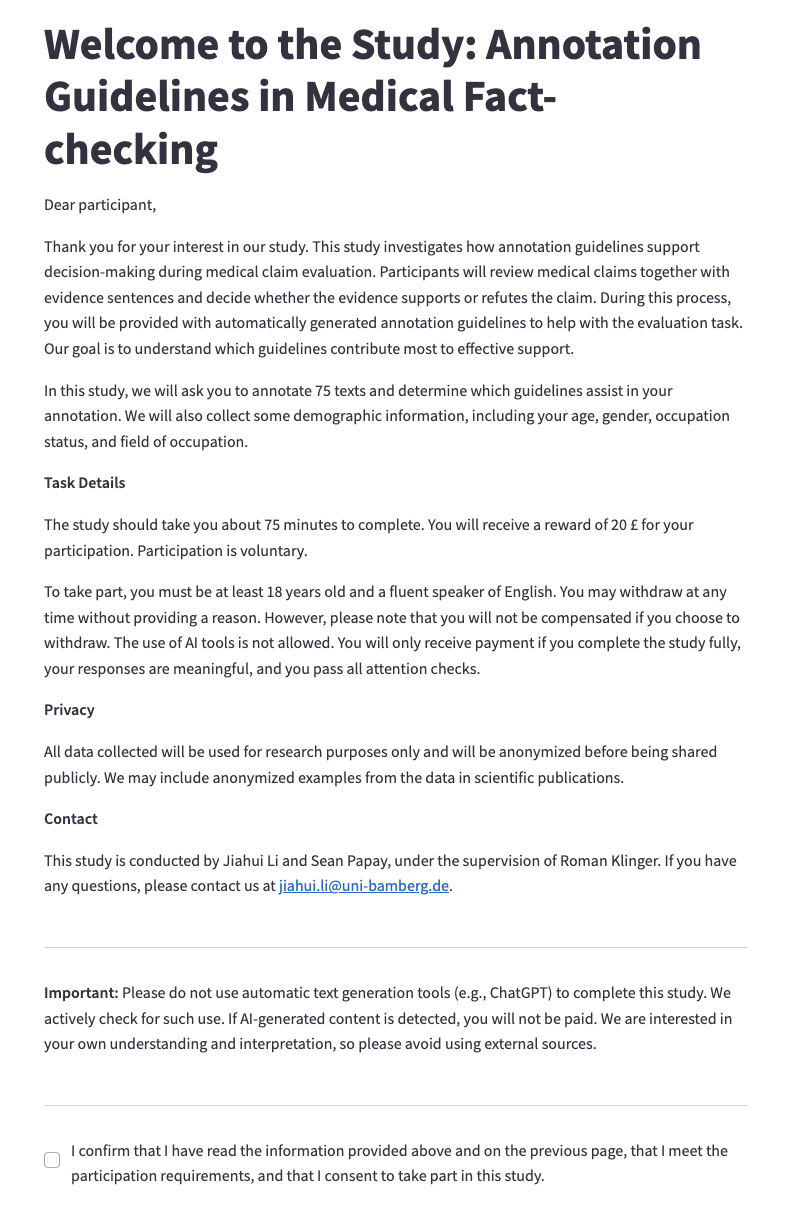} 
  \caption {A screenshot of the content page for the medical fact-checking survey.}
  \label{fig:screen-health}
\end{figure*}

\begin{figure*}[htbp]
\centering
  \includegraphics[width=0.8\linewidth]{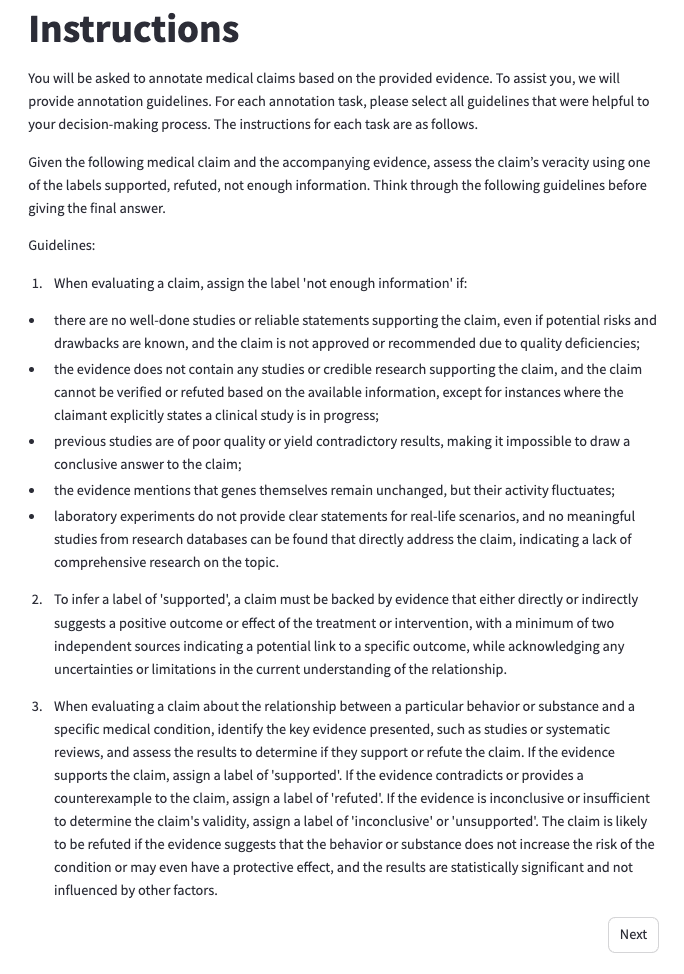} 
  \caption {A screenshot of the instruction page for the medical fact-checking survey.}
  \label{fig:screen-ins}
\end{figure*}

\begin{figure*}[htbp]
\centering
  \includegraphics[width=0.8\linewidth]{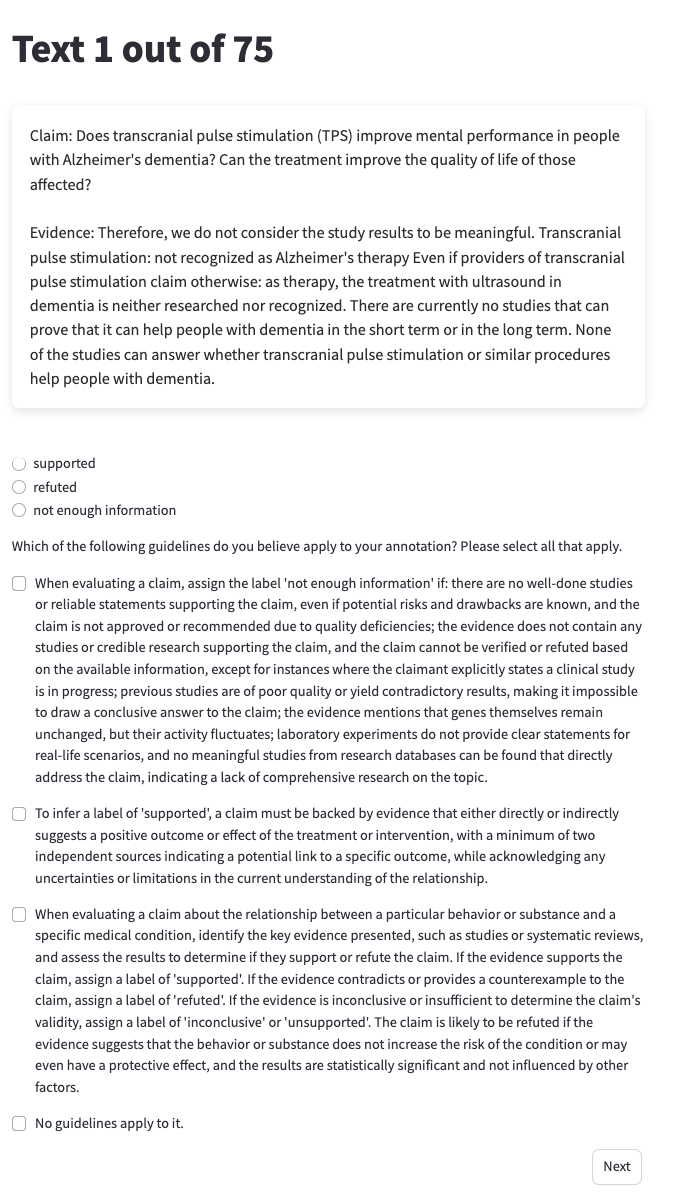} 
  \caption {A screenshot example of the annotation task for the medical fact-checking survey.}
  \label{fig:screen-text}
\end{figure*}

We provide the screenshots of the consent and instruction pages guiding our human study surveys. See Figure~\ref{fig:screen-health} and Figure~\ref{fig:screen-ins} for the medical fact-checking task as an example. An instance of the annotation case is shown in Figure~\ref{fig:screen-text}.

\clearpage
\onecolumn
\begin{center}
\small
\begin{longtable}{p{1cm}p{1.3cm}p{12.2cm}}
    \caption{Guidelines development in the optimization process for the HealthVer dataset with the iPOE-llm method using the Qwen3-4B-Instruct-2507 model.}
    \label{tab:gls} \\
   
    \toprule
        \textbf{Iteration} & \textbf{Operation} & \textbf{Guidelines}\\
    \midrule
    \endfirsthead

\multicolumn{3}{c}{{\tablename\ \thetable{} -- continued from previous page}} \\
\toprule
\textbf{Iteration} & \textbf{Operation} & \textbf{Guidelines} \\
\midrule
\endhead

\midrule
\multicolumn{3}{r}{{Continued on next page}} \\
\bottomrule
\endfoot

\bottomrule
\endlastfoot
        1 & add & 1. If the evidence explicitly mentions or lists a cardiovascular or cerebrovascular condition, such as hypertension or stroke, as a predictor or risk factor for severe COVID-19 illness, and the claim states that such conditions may increase the risk of severe illness from COVID-19, then the claim is supported. This includes cases where hypertension or other related conditions are ranked among the strongest or significant predictors of severe illness, as stated in the evidence, even if other conditions like diabetes or obesity are also mentioned. 2. If the evidence discusses immunological mechanisms such as acquired immunity or protection against re-exposure to a virus, and does not address the subjective experience, duration, or physical/mental sensations associated with recovery from a viral infection, then the relationship between the claim and evidence is neutral because the evidence neither supports nor contradicts the claim about the feeling of recovery. The label should be 'Neutral' when the evidence is scientifically relevant to immunity but irrelevant to the sensory or experiential aspects of recovery. 3. A claim that asserts both a supported effect (e.g., face masks reducing COVID-19 spread) and an unsupported or unmentioned effect (e.g., masks working against other pathogens) should be labeled 'not enough information' if the evidence only supports the first part and provides no information about the second. The label 'Supports' is only appropriate when the evidence directly and fully supports all components of the claim; otherwise, if any part of the claim is not addressed or cannot be verified by the evidence, the correct label is 'not enough information'.  \\
        4 & replace &  1. If a claim asserts a strong, specific effect such as'significantly improving' symptoms in a particular disease (e.g., COVID- for a treatment (e.g., Vitamin C), and the evidence only indicates potential benefit for respiratory viral infections or nutrient deficiency without specifying a significant improvement in symptoms for that specific condition, then the claim should be labeled as refuted because the strength and specificity of the claim exceed what is supported by the evidence. 2. If the evidence indicates that nicotine has a preventive or modulatory effect on an inflammatory response or other physiological process in the lungs, and this effect is presented as a plausible biological mechanism that could explain the protective effects of smoking, then the claim is supported. This includes cases where the evidence suggests a direct or indirect role of nicotine in reducing inflammation or modulating lung function, even in the context of smoking's overall harm. 3. If the evidence provides a direct, quantitative relationship showing that an increase in temperature leads to a reduction in the daily growth rate of COVID-19 cases, even if the reduction is modest, then the claim that 'COVID-19 cases drop in warm weather, but not much' is supported. This includes cases where temperature is identified as a key factor influencing case growth, with specific numerical data indicating a decrease in case rates per degree Celsius rise in temperature, regardless of the magnitude of the drop. \\
        10 & merge & 1. A claim is supported if the evidence provides a direct, quantitative, or plausible biological mechanism that aligns with the specificity and strength of the assertion—such as a demonstrated effect of temperature on case growth, a biological role of nicotine in modulating lung inflammation, or a specific, measurable benefit of a treatment on a defined condition. However, if the claim asserts a strong, specific effect (e.g., significantly improving symptoms of a particular disease) that is not directly or sufficiently supported by the evidence—especially when the evidence only indicates potential benefit in broader contexts—then the claim is refuted.  \\
        14 & merge & 1. A claim is supported if the evidence provides a direct, quantitative, or plausible biological mechanism that aligns with the specificity and strength of the assertion—such as a demonstrated effect of temperature on virus transmission, a biological role of a substance in modulating disease processes, or a specific, measurable benefit of a treatment on a defined condition. Claims of strong, specific effects—such as significantly improving symptoms of a particular disease—may be refuted if the evidence only suggests potential benefit in broader contexts. Additionally, claims that a vaccine developed for one condition offers non-specific protection against another disease—such as COVID-19—may be supported if the evidence indicates such protection could reduce transmission or save lives. Similarly, claims that specific climatological factors (like lower winter humidity) contribute to faster spread of a virus are considered supported when the evidence implies a relevant relationship between environmental conditions and viral transmission, even if not explicitly stated. \\
        18 & merge & 1. A claim is supported if the evidence provides a direct, quantitative, or plausible biological mechanism that aligns with the specificity and strength of the assertion—such as a demonstrated effect of temperature on virus transmission, a biological role of a substance in modulating disease processes, or a specific, measurable benefit of a treatment on a defined condition. Claims of strong, specific effects—such as significantly improving symptoms of a particular disease—may be refuted if the evidence only suggests potential benefit in broader contexts. Claims of non-specific protection, such as a vaccine developed for one condition offering protection against another disease like COVID-19, are supported if the evidence indicates such protection could reduce transmission or save lives. Similarly, claims linking climatological factors (e.g., lower winter humidity) to faster virus spread are supported when the evidence shows a relevant relationship between environmental conditions and transmission, even if not explicitly stated. However, a claim that a condition increases risk for complications and includes a mention of risk reduction methods should be labeled 'Supports' only if the evidence explicitly addresses both the increased risk and the availability or effectiveness of mitigation strategies; otherwise, it should be labeled 'not enough information' due to the lack of substantiation for risk reduction methods. \\
        34 & merge & 1. A claim is supported if the evidence provides a direct, quantitative, or plausible biological mechanism that aligns with the specificity and strength of the assertion—such as a demonstrated effect of temperature on virus transmission, a biological role of a substance in modulating disease processes, or a specific, measurable benefit of a treatment on a defined condition. Claims of strong, specific effects—such as significantly improving symptoms of a particular disease—may be refuted if the evidence only suggests potential benefit in broader contexts. Claims of non-specific protection, such as a vaccine developed for one condition offering protection against another disease like COVID-19, are supported if the evidence indicates such protection could reduce transmission or save lives. Similarly, claims linking climatological factors (e.g., lower winter humidity) to faster virus spread are supported when the evidence shows a relevant relationship between environmental conditions and transmission, even if not explicitly stated. However, a claim that a condition increases risk for complications and includes a mention of risk reduction methods should be labeled 'Supports' only if the evidence explicitly addresses both the increased risk and the availability or effectiveness of mitigation strategies; otherwise, it should be labeled 'not enough information' due to the lack of substantiation for risk reduction methods. \\
        39 & merge & 1. A claim is supported if the evidence provides a direct, quantitative, or plausible biological mechanism that aligns with the specificity and strength of the assertion—such as a demonstrated effect of temperature on virus transmission, a biological role of a substance in modulating disease processes, or a specific, measurable benefit of a treatment on a defined condition. Claims of strong, specific effects—such as significantly improving symptoms of a particular disease—may be refuted if the evidence only suggests potential benefit in broader contexts. Claims of non-specific protection, such as a vaccine developed for one condition offering protection against another disease like COVID-19, are supported if the evidence indicates such protection could reduce transmission or save lives. Similarly, claims linking climatological factors (e.g., lower winter humidity) to faster virus spread are supported when the evidence shows a relevant relationship between environmental conditions and transmission, even if not explicitly stated. However, a claim that a condition increases risk for complications and includes a mention of risk reduction methods should be labeled 'Supports' only if the evidence explicitly addresses both the increased risk and the availability or effectiveness of mitigation strategies; otherwise, it should be labeled 'not enough information' due to the lack of substantiation for risk reduction methods. \\
        43 & replace & 1. A claim that a face mask will protect you from Covid-19 is supported if the evidence acknowledges or aligns with the medical understanding that masks can reduce transmission by blocking respiratory droplets, even if it notes limited usage or societal adoption. The label'refutes' should only be applied when the evidence directly contradicts the protective efficacy of face masks, such as by claiming they are ineffective or have no impact on transmission. If the evidence merely points out low usage without challenging the protective function, the claim is supported and the label should be'supported' rather than'refutes'. \\
        235 & merge & 1. A claim that a face mask will protect you from Covid-19 is supported if the evidence acknowledges or aligns with the medical understanding that masks can reduce transmission by blocking respiratory droplets, even if it notes limited usage or societal adoption. The label'refutes' should only be applied when the evidence directly contradicts the protective efficacy of face masks, such as by claiming they are ineffective or have no impact on transmission. If the evidence merely points out low usage without challenging the protective function, the claim is supported and the label should be'supported' rather than'refutes'. \\
        316 & add & 1. A claim that a face mask will protect you from Covid-19 is supported if the evidence acknowledges or aligns with the medical understanding that masks can reduce transmission by blocking respiratory droplets, even if it notes limited usage or societal adoption. The label'refutes' should only be applied when the evidence directly contradicts the protective efficacy of face masks, such as by claiming they are ineffective or have no impact on transmission. If the evidence merely points out low usage without challenging the protective function, the claim is supported and the label should be'supported' rather than'refutes'. 2. If the evidence states there is no supporting evidence to discourage the use of a medication (such as ibuprofen) in the context of symptoms like fever and COVID-19, the claim that the medication should be avoided because it can worsen the disease is refuted, as the evidence directly contradicts the claim's validity by lacking any scientific basis to support the recommended avoidance. 3. If the evidence only describes the methods or tools used to assess mental health without providing any specific findings, data, or results related to the populations mentioned in the claim (such as youth, minorities, or essential workers), the label should be 'Neutral' because the evidence does not support, contradict, or verify the claim; the absence of specific findings means the claim cannot be evaluated based on the provided information. 4. Label 'Supports' when the evidence confirms or aligns with the claim regarding the origin of the virus (e.g., bat origin) and either provides indirect support through genomic clustering with bat or pangolin coronaviruses, while acknowledging that the specific intermediate animal transmitting the virus to humans remains unknown, as stated in the claim. The label should be assigned if the evidence does not contradict the claim and maintains consistency with the stated scientific uncertainty about the transmission pathway. \\
        
\end{longtable}
\end{center}
\twocolumn


\end{document}